\documentclass{article}

\usepackage{arxiv}

\usepackage[utf8]{inputenc} 
\usepackage[T1]{fontenc}    
\usepackage{hyperref}       
\usepackage{url}            
\usepackage{booktabs}       
\usepackage{amsfonts}       
\usepackage{nicefrac}       
\usepackage{microtype}      
\usepackage{lipsum}		
\usepackage{graphicx}
\usepackage{natbib}
\usepackage{doi}

\usepackage[linesnumbered,ruled,vlined]{algorithm2e}
\usepackage{graphicx}

\usepackage{times}
\usepackage{latexsym}

\title{Applying SoftTriple Loss for Supervised Language Model Fine Tuning}

\author{ \href{https://orcid.org/0000-0002-2241-9588}{\hspace{1mm}Witold Sosnowski} \\
	Faculty of Mathematics and Information Science\\
	Warsaw University of Technology\\
	Warsaw, Poland \\
	\texttt{witold.sosnowski.dokt@pw.edu.pl} \\
	\And
	\href{https://orcid.org/0000-0002-3407-7570}{\hspace{1mm}Anna Wróblewska} \\
	Faculty of Mathematics and Information Science\\
	Warsaw University of Technology\\
	Warsaw, Poland \\
	\texttt{anna.wroblewska1@pw.edu.pl} \\
	\And
	\href{https://orcid.org/0000-0002-9647-6761}{\hspace{1mm}Piotr Gawrysiak} \\
	Faculty of Electronics and Information Technology\\
	Warsaw University of Technology\\
	Warsaw, Poland \\
	\texttt{p.gawrysiak@ii.pw.edu.pl} \\
}

\renewcommand{\shorttitle}{SoftTriple Loss for Supervised Language Model Fine Tuning}

\hypersetup{
pdftitle={A template for the arxiv style},
pdfsubject={q-bio.NC, q-bio.QM},
pdfauthor={David S.~Hippocampus, Elias D.~Striatum},
pdfkeywords={First keyword, Second keyword, More},
}

\begin{document}
\maketitle







\begin{abstract}
We introduce a new loss function TripleEntropy to improve classification performance for fine-tuning general knowledge pre-trained language models based on cross-entropy and SoftTriple loss.
This loss function can improve the robust RoBERTa baseline model fine-tuned with cross-entropy loss by about (0.02\% - 2.29\%).
Thorough tests on popular datasets indicate a steady gain. The fewer samples in the training dataset, the higher gain -- thus, for small-sized dataset it is 0.78\%, for medium-sized -- 0.86\% for large -- 0.20\% and for extra-large 0.04\%.

\end{abstract}

\section{Introduction}
Natural language processing (NLP) is a rapidly growing area of machine learning with applications wherever a computer needs to operate on a text that involves capturing its semantics. It may include text classification, translation, text summarization, question answering, dialogues. All these tasks are upstream and depend on the quality of the text representation~\cite{white2015well}. Many models can produce such text representations, from Bag-Of-Word or Word2Vec word embedding to the state-of-the-art language representation model BERT with variations in most NLP tasks. 

The best performance on text classification tasks is obtained when the model is first trained on a general knowledge corpus to capture semantic relationships between words and then fine-tuned with an additional dense layer on a domain corpus with cross-entropy loss \cite{radford2019language}.

We introduce a new loss function TripleEntropy to improve classification performance for fine-tuning general knowledge pre-trained language models based on cross-entropy loss and SoftTriple loss \cite{devlin2018bert, qian2019softtriple}. Triplet Loss transforms the embedding space so that vector representations from the same class can form separable subspaces, stabilizing, and generalizing the language model fine-tuning process. TripleEntropy can improve the fine-tuning process of the RoBERTa based models so the performance on downstream task increases by about (0.02\% - 2.29\%).

In the following sections, we review relevant work on state-of-the-art in distance metric learning (Section~\ref{sec:related-work}); describe our approach for training and our metric SoftTriple loss and  outline the experimental setup (Section~\ref{sec:our-approach}); discuss the results (Section~\ref{sec:results}); conclude and offer directions for further research (Section~\ref{sec:conclusions}). 

\section{Related Work}\label{sec:related-work}

\subsection{Building Sentence Embeddings}
Building embeddings that represent sentences is challenging because the natural language can be very diverse. The meaning can change drastically depending on the context of a word. It is also an important issue because the quality of sentence embeddings substantially impacts the performance of all downstream tasks like text classification and question answering. Because of that, so far, considerable research effort has been put into building sentence embeddings. 

One of the first vector representations (embeddings), bag-of-words (BOW), is an intriguing approach in which the text is represented as a bag (multiset) of its words, with each word represented by its occurrence in the text~\cite{parsing2009speech}. The disadvantage of this strategy was that the embeddings were handcrafted, unlike the Word2Vec approach, which used a machine learning process to predict word embeddings~\cite{mikolov2013efficient}. In Word2Vec, each word embedding is selected based on its overall context in the training corpus and can express the latent semantic of words. It automatically expresses the semantics of the whole sentences, though, so several approaches were proposed to tackle this problem. The most popular was representing the sentence embedding as a weighted average of the sentence's word vectors. Because every word has the same embedding regardless of its meaning in the entire sentence, such an approach is not resistant to sentence changes and context semantics. 

Bidirectional Encoder Representations from Transformers (BERT) is a very well known technique for constructing high-quality sentence embeddings that can express the dynamic and latent meaning of the whole sentences better than any previous approach. Its sentence embeddings can accurately reflect the meaning of the input text, making a significant difference in the quality of the downstream tasks performed. An even better variant of the BERT-based architecture, RoBERTa, has emerged and has lately become unquestionably state-of-the-art in terms of sentence embedding construction \cite{liu2019roberta, dadas2020pre}.

\subsection{Distance Metric Learning}

Learning embeddings where instances from the same class are closer than examples from other classes is known as Distance Metric Learning (DML)~\cite{qian2019softtriple}. DML recently has drawn much attention due to its wide applications, especially in image processing. It can be used in the classification tasks together with the k-nearest neighbour algorithm \cite{weinberger2009distance}, clustering along with K-means algorithm \cite{xing2002distance} and semi-supervised learning \cite{wu2020metric}. DML's objective is to create embeddings similar to examples from the same class but different from observations from other classes. \cite{movshovitz2017no}. In contrary to the cross-entropy loss, which only takes care of intra-class distances to make them linearly separable, the DML approach maximizes inter-class and minimizes the intra-class distances \cite{wen2016discriminative}. Aside from that, a typical classifier based solely on cross-entropy loss concentrates on class-specific characteristics rather than generic ones, as it is only concerned with distinguishing between classes rather than learning their representations. DML focuses on learning class representations, making the model more generalizable to new observations and more robust to outliers.  

\subsubsection{Contrastive Loss}
Contrastive Loss is one of the methods in DML \cite{hadsell2006dimensionality}. It concentrates on pairs of similar and dissimilar observations, whose distances are attempted to be minimized if they belong to the same class and maximized if they belong to different classes. The loss function is given in Equation~\ref{eq:contrastive-loss}.

\begin{equation} \label{eq:contrastive-loss}
L\left(W,\left(Y, \overrightarrow{X}_{1}, \overrightarrow{X_{2}}\right)^{i}\right)=(1-Y) L_{S}\left(D_{W}^{i}\right)+Y L_{D}\left(D_{W}^{i}\right)
\end{equation}

\noindent where $\left(Y, \overrightarrow{X}_{1}, \overrightarrow{X}_{2}\right)^{i}$ denotes the labeled sample pair of with the index $i$, $L_{S}$ represents the loss function for a pair of similar points, $L_{D}$ is the loss function applied for pair of dissimilar points and $D_{W}$ denotes distance function between pair of points $\overrightarrow{X_{1}}, \overrightarrow{X_{2}}$.

\subsubsection{Triplet Loss} 
Triplet Loss is similar to Contrastive Loss but works with triplets instead of pairs, is another solution to the DML problem \cite{schroff2015facenet}. Each triplet comprises an anchor, a positive, and a negative observation. Positive examples are members of the same class as an anchor, but negative instances belong to a separate class. Because it considers more observation simultaneously, it optimizes the embedding space better than Contrastive Loss. The actual formula for Triplet Loss is in Equation~\ref{eq:triplet-loss}.

\begin{equation}\label{eq:triplet-loss}
L=\sum_{i=1}^{N}\left[\left\|f_{i}^{a}-f_{i}^{p}\right\|_{2}^{2}-\left\|f_{i}^{a}-f_{i}^{n}\right\|_{2}^{2}+\alpha\right]_{+}
\end{equation}

\noindent where $f(x)$ represents the embedding that embeds ab observation $x$ into a $d$-dimensional Euclidean space. $x_{i}^{a}$ denotes an $anchor$, $x_{i}^{p}$ $(positive)$ is the observation from the same class as the anchor, $x_{i}^{n}$ $(negative)$ denotes an observation belonging to a different than the anchor class, $\alpha$ is an imposed between positive and negative pairs margin. 

The most typical issue with triplets and contrastive learning is that as the number of observations in a batch grows, the number of pairs and triplets grows squarely or cubically. Another issue that might arise is the use of training pairs and triplets that are relatively easy to distinguish, leading to poor model generalization. Semi-solutions of the above problems are as introducing $\tau$ a temperature parameter that controls the separation of classes \cite{chen2020simple}, or hard triples, which creates triplets based on harder negatives \cite{hermans2017defense}. 

\subsubsection{ProxyNCA Loss}
It is a more general approach to solving a problem with high resource consumption \cite{movshovitz2017no}. It employs proxies, artificial data points that represent the entire dataset. One proxy approximates one class; therefore, there are as many proxies as classes. This technique drastically reduces the number of triplets while simultaneously raising the convergence rate since each proxy make the triplet more resistant to outliers. The proxies are integrated into the model as trainable parameters since synthetic data points are represented as embeddings. Equation~\ref{eq:proxynca-loss} depicts a ProxyNCA loss formula.

\begin{equation} \label{eq:proxynca-loss}
L=-\log \left(\frac{\exp \left(-d\left(\frac{x_{i}}{\| x_{i}||_{2}}, \frac{f\left(x_{i}\right)}{\left\|f\left(x_{i}\right)\right\|_{2}}\right)\right)}{\sum_{f(z) \in Z} \exp \left(-d\left(\frac{x_{i}}{\left\|x_{i}\right\|_{2}}, \frac{f(z)}{\|f(z)\|_{2}}\right)\right)}\right)
\end{equation}

\noindent where $C_{i}$ is a set of observations from the same class, $f(a)$ denotes a proxy function returning class proxy for given parameter $a$, $\left.|a|\right|_{2}$ is the $L^{2}$-Norm of the vector $a$, $d\left(x_{i}, f\left(x_{i}\right)\right)$ denotes a distance between the sample $x_{i}$ and proxy $f\left(x_{i}\right)$, $Z$ denotes set of all proxies, where 
$f(z) \in Z$ and $z \notin C_{i}$.

A single proxy per class may not be enough to represent the class's inherent structure in real-world data. Another DML loss function has been created that introduces multiple proxies per class - SoftTriple Loss \cite{qian2019softtriple}. ProxyNCA Loss can produce better embeddings while maintaining a smaller number of triplets than Triplet Loss or Contrastive Loss. The SoftTriple Loss is defined by the formulas in Equations~\ref{eq:softtriple-loss} and \ref{eq:softtriple-loss2}.

\begin{equation} \label{eq:softtriple-loss}
\ell_{SoftTriple}
=-\log \frac{\exp \left(\lambda\left(\mathcal{S}_{i, y_{i}}^{\prime}-\delta\right)\right)}{\exp \left(\lambda\left(\mathcal{S}_{i, y_{i}}^{\prime}-\delta\right)\right)+\sum_{j \neq y_{i}} \exp \left(\lambda \mathcal{S}_{i, j}^{\prime}\right)}
\end{equation}

\begin{equation} \label{eq:softtriple-loss2}
\mathcal{S}_{i, c}^{\prime}=\sum_{k} \frac{\exp \left(\frac{1}{\gamma} \mathbf{E({x}_{i})}^{\top} \mathbf{w}_{c}^{k}\right)}{\sum_{k} \exp \left(\frac{1}{\gamma} \mathbf{E({x}_{i})}^{\top} \mathbf{w}_{c}^{k}\right)} \mathbf{E({x}_{i})}^{\top} \mathbf{w}_{c}^{k}
\end{equation}

\noindent where, $C$ denotes the class number, $k$ is the number of proxies representing observations from SoftTriple for each class, $\delta$ defines a margin between the example and class centers from different classes, $\lambda$ reduces the influence from outliers and makes the loss more robust, $\gamma$ is the scaling factor for the entropy regularizer, $x_{i}$ defines the single observation represented as an array of tokens, $E(\cdot) \in \mathbf{R}^{d}$ indicates the encoder, $\mathbf{w}_{c}^{k}$ are weights representing proxy embeddings of the class $c$ (there are $k$ of them).

\section{Our Approach} \label{sec:our-approach}

For fine-tuning pre-trained language models, we offer a novel objective function. It is based on the supervised cross-entropy loss and the SoftTriple Loss \cite{qian2019softtriple}. The latter component is a loss from the Distance Metric Learning (DML) family of losses, which learns an embedding by capturing similarities between embeddings from the same class and distinguishing them from embeddings from different classes \cite{qian2019softtriple}.

For a classification problem let us denote: 
\begin{itemize}
    \item $N$ the number of observations,
     \item $C$  the class number,
    \item $y_{i c}$ the objective probability of the class $c$ for the $i$th observation,
    \item $\beta$ the scaling factor that tunes influence of both parts of the loss.
\end{itemize}  
The novel goal function is given by the following formula:
\begin{equation}
\mathcal{L}=(\beta) \mathcal{\ell}_{MC E}+(1 - \beta) \mathcal{\ell}_{SoftTriple}
\label{eq:novel-loss}
\end{equation}
, where
\begin{equation}
\ell_{{MCE }} \\
=-\frac{1}{N} \sum_{i}^{N} \sum_{c}^{C} y_{i c} \log \left(p_{i c}\right)
\end{equation}

It can be applied for different encoders $E(\cdot) \in \mathbf{R}^{d}$ from both image and natural language processing domains.

\subsection{Model}

In our work, we use the objective function from Equation \ref{eq:novel-loss} to fine-tune the pre-trained BERT-based language models provided by the \textit{huggingface} library as RoBERTa-base and RoBERTa-large. In the standard setting, the single input text is first tokenized with WordPiece embeddings \cite{wu2016google}, which produces a vector of tokens $x_{i}$ with a maximum length of 512, with $[CLS]$ at the beginning of an array, $[EOS]$ at the end and $[SEP]$ between tokens representing different sentences. The output of RoBERTa model $E({x_{i}}) \in \mathbf{R}^{d}$ is an array of embeddings, where each input token has its corresponding embedding. 

\subsubsection{Multinominal Cross-Entropy Loss} 
In our experiments, we used the multinominal cross-entropy loss calculated in the same way as it was proposed by the authors of the BERT language model\cite{devlin2018bert}. The sentence representation is obtained by pooling the output of the model  $E({x_{i}}) \in \mathbf{R}^{d}$ and passing it to the $C$ dimensional single fully connected layer. Its output is passed to the softmax function generating probabilities $p_{i c}$, which are, along with objective probabilities $y_{i c}$, directly feeding the multinominal cross-entropy loss.

\subsubsection{SoftTriple Loss}
The second component of the TripleEntropy \ref{eq:novel-loss} is SoftTriple Loss (3), responsible for a more robust and better generalization of the model during tuning. It is fed by the direct output of the model $E({x_{i}}) \in \mathbf{R}^{d}$, even before pooling. It means that if the batch size is $B$, then the total number of embeddings that feed SoftTriple Loss during one training iteration is $B$ * $|x_{i}|$. This implementation ensures that the proxies representing each class will be well approximated so that the quality of fine-tuning increases.
  

Our implementation is a development of the earlier work \cite{gunel2020supervised}, where Contrastive Loss was applied only to the embedding corresponding to the first $[CLS]$ token of the input vector $x_{i}$. We apply SoftTriple Loss to the embeddings corresponding to all tokens from the input vector $x_{i}$, which ensures the better generalization of the fine-tuning process but requires more computing power.
Fortunately, the SoftTriple Loss is significantly more efficient than the Contrastive Loss since it generates triplets not from all observations but its approximated proxies.


\subsection{Training Procedure} \label{training_proc}

Each result (average accuracy) was obtained as based on 4 seed runs (2, 16, 128, 2048), where each run was 5-fold cross-validated. It means that each accuracy result is an averaged of 20 different results. Apart from that, each result was based on the best parameter combination obtained by grid search which included parameters $k \in\{10, 100, 1000\}$, $\gamma \in\{0.01, 0.03, 0.05, 0.07, 0.1\}$, $\lambda \in\{1,3,3.3,4,6,8,10\}$, $\delta \in\{0.01, 0.1, 0.3, 0.5, 0.7, 0.9, 1\}$ and $\beta \in\{0.1, 0.3, 0.5, 0.7, 0.9\}$


\subsection{Datasets}
We employed a variety of well-known datasets from SentEval \cite{conneau2018senteval} along with the IMDb \cite{maas2011learning} for model evaluations that covered both text classification and textual entailment as two important natural language tasks in order to assess the general use of TripleEntropy. Table~\ref{tab:datasets_description} shows the description of the datasets.

\begin{table}[h]
\centering
\begin{tabular}{|l|l|l|l|}
\hline
Dataset & \# Sentences & \# Classes & Task\\
\hline
SST2 & 67k & 2 & Sentiment (movie reviews)\cite{socher2013recursive}\\
IMDb & 50k & 2 & Sentiment (movie reviews) \cite{maas2011learning}\\
MR & 11k & 2 & Sentiment (movie reviews) \cite{pang2005seeing} \\
MPQA & 11k & 2 & Opinion polarity \cite{wiebe2005annotating}\\
SUBJ & 10k & 2 & Subjectivity status \cite{pang2004sentimental}\\
TREC & 5k & 6 & Question-type classification \cite{pang2005seeing}\\
CR & 4k & 2 & Sentiment (product review) \cite{hu2004mining}\\
MRPC & 4k & 2 & Paraphrase detection \cite{dolan2004unsupervised}\\
\hline
\end{tabular}
\caption{SentEval and IMDb datasets used for our evaluation.}
\label{tab:datasets_description}
\end{table}

Additionally, we have examined the performance of our method when the number of training examples is limited to 1,000 and 10,000 observations on sampled datasets according to Table~\ref{tab:few_shot_datasets_description}.

\begin{table}[h]
\centering
\begin{tabular}{|l|l|l|l|}
\hline
Dataset & \# Sentences & \# Casses & Description\\
\hline
IMDb-10k & 10k & 2 & Sampled 10k subset from sentiment (movie reviews) \cite{maas2011learning}\\
SST2-10k & 10k & 2 & Sampled 10k subset from sentiment (movie reviews)\cite{socher2013recursive}\\
MR-1k & 1K & 2 & Sampled 1k subset from sentiment (movie review) \cite{pang2005seeing}\\
CR-1k & 1K & 2 & Sampled 1k subset from sentiment (product review) \cite{hu2004mining}\\
TREC-1k & 1K & 6 & Sampled 1k subset from question-type classification \cite{pang2005seeing}\\
MRPC-1k & 1K & 2 & Sampled 1k subset from paraphrase detection \cite{dolan2004unsupervised}\\
IMDb-1k & 1K & 2 & Sampled 1k subset from sentiment (movie review) \cite{maas2011learning}\\
SST2-1k & 1K & 2 & Sampled 1k subset from sentiment (movie review) \cite{socher2013recursive}\\
MPQA-1k & 1K & 2 & Sampled 1k subset from opinion polarity \cite{wiebe2005annotating}\\
SUBJ-1k & 1K & 2 & Sampled 1k subset from subjectivity status \cite{pang2004sentimental}\\
\hline
\end{tabular}
\caption{Sampled SentEval and IMDb datasets used for our evaluation.}
\label{tab:few_shot_datasets_description}
\end{table}

\section{Results} \label{sec:results}

Results are presented in the form of comparison between the performance of the RoBERTa-base (RB) and the RoBERTa-large (RL) models as a baselines and the RoBERTa-base with SoftTriple Loss (RB SoftTriple) as well as RoBERTa-large with SoftTriple Loss (RL SoftTriple). Moreover, we have created 4 groups depending on the size of the dataset. In the first group, we present results regarding the small-sized datasets with the number of sentences of 1,000. In the second group, we explore results for the medium-sized datasets in which the number of sentences is not greater than 5,000 and not smaller than 4,000. In the third group, we present results belonging to the large-sized datasets with the number of sentences larger than 10,000 and fewer than 11,000. The extra-large-sized group consists of elements where the number of observations is larger than 50,000.  

The RB baseline models were trained with the use of AdamW optimizer \cite{kingma2014adam}, beginning learning rate 1e-5, L2 regularization, learning rate scheduler and linear warmup from 0 to 1e-5 for the first 6\% of steps and batch size of 64. The RB SoftTriple models were trained on the same set of hyperparameters as the baseline models they refer to and additional parameters specific to SoftTriple Loss as it is described in Section~\ref{training_proc}.

\subsection{RB SoftTriple for small datasets}
Table~\ref{tab:corpora-rb-small} presents the results for the datasets containing 1,000 sentences. We observe that models trained using TripleEntropy have a higher performance than the baselines by about 0.78\% on average. It is worth noting that the gain in performance is observed at each dataset, especially for the TREC-1k and MRPC-1k, where it amounts to 2.29\% and 1.11\%, respectively.

\begin{table}[h]
\centering
\begin{tabular}{|l|l|l|l|l|l|l|l|l|l|l|l|}
\hline
Model & SST2-1k  & IMDb-1k & SUBJ-1k & MPQA-1k & MRPC-1k & TREC-1k & CR-1k & MR-1k \\
\hline
RB & 88.63 & 81.00  & 94.61 & 87.75 & 78.01 & 79.80 & 91.57& 85.89
\\
RB SoftTriple & 89.09 & 81.45 & 94.70 & 87.93 & 79.12 & 82.09 & 92.16 & 86.39 \\
\hline
\end{tabular}
\caption{RoBERTa-base (RB) vs RoBERTa-base with SoftTriple Loss for small sized datasets}
\label{tab:corpora-rb-small}
\end{table}

\subsection{RB SoftTriple for medium datasets}
Table~\ref{tab:corpora-rb-medium} shows the results based on the datasets containing more than 1,000 sentences and less than 11,000. Here, we can observe that models trained using TripleEntropy have higher performance than the baselines by about 0,86\% on average. The highest gain in performance is observed in the case of TREC and MRPC datasets by 1.00\% and 1.28\%, respectively.

\begin{table}[h]
\centering
\begin{tabular}{|l|l|l|l|l|l|l|l|l|l|l|l|}
\hline
Model &  MRPC & TREC & CR & MR \\
\hline
RB &  83.11 & 96.19 & 93.28 & 89.09 \\
RB SoftTriple  & 84.39 & 97.19 & 93.58 & 89.29 \\
\hline
\end{tabular}
\caption{RoBERTa-base (RB) vs RoBERTa-base with SoftTriple Loss for medium sized datasets}
\label{tab:corpora-rb-medium}
\end{table}

\subsection{RB SoftTriple for large datasets}
Table~\ref{tab:corpora-rb-large} shows the results based on the datasets containing 10,000-11,000 sentences. The gain in the performance amounts 0.20\%. 

\begin{table}[h]
\centering
\begin{tabular}{|l|l|l|l|l|}
\hline
Model &  SST2-10k  & IMDb-10k &  SUBJ & MPQA\\
\hline
RB & 92.63 & 85.12 & 96.83 & 91.08
\\
RB SoftTriple & 92.79 & 85.23 & 97.15 & 91.30
\\
\hline
\end{tabular}
\caption{RoBERTa-base (RB) vs RoBERTa-base with SoftTriple Loss for large sized datasets}
\label{tab:corpora-rb-large}
\end{table}

\subsection{RB SoftTriple for extra-large datasets}
Table~\ref{tab:corpora-rb-extra-large} shows the results based on the datasets containing more than 50,000 sentences. The gain in the performance is not as high as in the case of the medium and small-sized datasets, and it is 0.04\% on average, which is not significant. 

\begin{table}[h]
\centering
\begin{tabular}{|l|l|l|}
\hline
Model & SST2  & IMDb  \\
\hline
RB & 94.89 & 87.10 \\
RB SoftTriple & 94.95 & 87.12 \\
\hline
\end{tabular}
\caption{RoBERTa-base (RB) vs RoBERTa-base with SoftTriple Loss for extra large sized datasets}
\label{tab:corpora-rb-extra-large}
\end{table}

\subsection{RL SoftTriple for small datasets}
We have compared our results to the related work \cite{gunel2020supervised} where the authors claim the performance gains over baseline RoBERTa-large by applying loss function consisted of cross-entropy loss and Supervised Contrastive Learning loss. The work shows the improvement over baseline in the few-shot learning defined as fine-tuning based on the training dataset consisted of 20, 100 and 1,000 observations. In order to compare our new loss function with the results from the related work we conducted experiments where the baseline was RoBERTa-large (RL) with cross-entropy loss and compared it to the RoBERTa-large with cross-entropy and SoftTriple loss (RL SoftTriple) on the dataset consisted of 1,000 observations. We can observe that our method yields a gain over baseline of 0.48\%, which is higher than the performance improvement over baseline for a dataset of the same size from related work, whose improvement over baseline is 0.27\%. The results are presented on the table \ref{tab:corpora-rl-small}.

\begin{table}[h]
\centering
\begin{tabular}{|l|l|l|l|l|l|l|l|l|l|l|}
\hline
Model & SST2-1k & MPQA-1k & MRPC-1k & TREC-1k & CR-1k & MR-1k  \\
\hline
RL & 91.96  & 90.18 &  76.09 & 83.75  & 93.43 & 89.69
\\
RL SoftTriple & 92.14 &  90.59 & 77.16& 84.59  & 93.62 & 89.89
\\
\hline
\end{tabular}
\caption{RoBERTa-large (RB) vs RoBERTa-large with SoftTriple Loss for small sized datasets}
\label{tab:corpora-rl-small}
\end{table}

\subsection{Discussion}
We can observe that our method improves the performance most significantly for the small-sized dataset by 0.87\% in the case of RoBERTa-base baseline and 0.48\% in the case of RoBERTa-large baseline and the medium-sized dataset, where the increase amounts to 0.86\%. For the large-sized dataset, the increase over baseline is 0.20\%, while for the extra-large-sized dataset, the gain over baseline amounts 0.04\%. Our experiments show consistent performance improvement over baseline when using SoftTriple loss, which is highest for the small and medium-sized datasets and decreases for the large and extra-large sized datasets. It is a significant improvement over previous related work, where the performance improvement for the supervised classification tasks was achieved only for the few-shot learning settings~\cite{gunel2020supervised}. 

\begin{figure}[h]
    \centering
    \includegraphics[scale = 0.3]{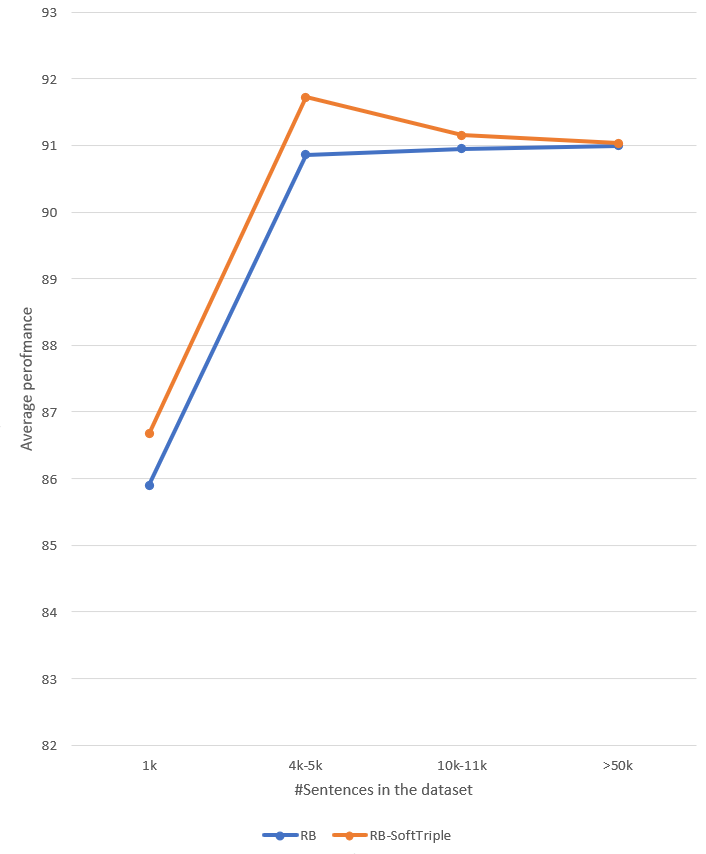}
    \caption{Performance comparison between RB and RB-SoftTriple}\label{fig:performance_comparison}
\end{figure}

We also conclude that the smaller the dataset is, the higher our new goal function's performance gain over baseline. The performance comparison between baseline and our method throughout dataset size is depicted in Figure~\ref{fig:performance_comparison}. 

\section{Conclusions} \label{sec:conclusions}

We proposed a supervised Distance Learning Metric objective that increases the performance of the RoBERTa-base models, which are strong baselines in the Natural Language Processing tasks. The performance is proved over multiple tasks from the single sentence classification and pair sentence classification to be higher by about (0.02\%-2.29\%) depending on the training dataset size. In addition, each result has been confirmed through tests with 5-fold cross-validation on 4 different seeds to increase its reliability. In the future, we plan to extend the application of our method to compare the results with language models from different architectures to investigate its general usefulness in other tasks.
\bibliography{bibliografia}
\bibliographystyle{acl_natbib}

\end{document}